%% file: acl_latex.tex
\pdfoutput=1

\PassOptionsToPackage{table,dvipsnames}{xcolor}

\documentclass[11pt]{article}

\usepackage[final]{acl}

\usepackage{times}
\usepackage{latexsym}
\usepackage[T1]{fontenc}
\usepackage[utf8]{inputenc}
\usepackage{microtype}
\usepackage{inconsolata}
\usepackage{graphicx}
\usepackage{multicol}
\usepackage{algpseudocode} 
\usepackage{tcolorbox}
\tcbuselibrary{skins, breakable, theorems}
\usepackage{amsmath}
\usepackage{amssymb}
\usepackage{amsthm}
\usepackage{booktabs}
\usepackage{framed} 
\usepackage{multirow}
\usepackage{enumitem}
\usepackage[compact]{titlesec}
\usepackage{float} 
\usepackage{etoolbox}
\usepackage{xspace}
\usepackage{accents}
\usepackage{subfig}
\usepackage{colortbl}
\usepackage{arydshln}
\usepackage{hyperref}
\usepackage[ruled,vlined]{algorithm2e} 
\usepackage{array}

\newcommand{\eat}[1]{}

\usepackage{adjustbox}

\definecolor{promptred}{HTML}{D45C47}
\definecolor{responseblue}{HTML}{3B61A6}
\definecolor{truthorange}{HTML}{D7A938}

\usepackage[normalem]{ulem}

\usepackage{bbding}
\usepackage{listings}
\usepackage{wrapfig}  %
\definecolor{bittersweet}{rgb}{1.0, 0.44, 0.37}
\definecolor{mygreen}{rgb}{0.29, 0.7, 0.48}
\usepackage{pifont}

\definecolor{demphcolor}{RGB}{144,144,144}

\definecolor{mygray}{gray}{0.4}
\definecolor{autopurple}{HTML}{7030A0}
\definecolor{dyna_yellow}{HTML}{BF8800}
\definecolor{adaptive_blue}{HTML}{0070C0}
\definecolor{darksalmon}{rgb}{0.91, 0.59, 0.48}
\definecolor{emerald}{rgb}{0.31, 0.78, 0.47}
\definecolor{green(pigment)}{rgb}{0.0, 0.65, 0.31}
\definecolor{amaranth}{rgb}{0.9, 0.17, 0.31}
\definecolor{iris}{rgb}{0.35, 0.31, 0.81}
\definecolor{uu}{rgb}{0.95, 0.51, 0.51}
\definecolor{spirodiscoball}{rgb}{0.06, 0.75, 0.99}
\usepackage{svg}

\usepackage{cleveref}
\SetKwInOut{Input}{Input}\SetKwInOut{Output}{Output}

\SetCommentSty{mycommfont}
\SetKwComment{Comment}{$\triangleright$\ }{}

\newcommand{\ourmethod}{{\fontfamily{lmtt}\selectfont \textbf{DS-Lighting}}\xspace}
\newcommand{\llmname}[1]{{\fontfamily{pcr}\selectfont {#1}}\xspace}

\definecolor{ada_blue}{rgb}{0,205,205}
\definecolor{glt_red}{rgb}{109,205,255}
\definecolor{MorandiBlue}{RGB}{118,134,146}

\definecolor{darkgrey}{RGB}{120,120,120}
\definecolor{mygrey}{RGB}{200,200,200}

\usepackage{makecell}
\usepackage{tabulary}

\definecolor{myblue}{HTML}{00CDCD}
\definecolor{champagne}{rgb}{0.74, 0.83, 0.9}
\definecolor{champagne}{rgb}{0.97, 0.91, 0.81}

\hypersetup{
    colorlinks=true,
    linkcolor=adaptive_blue,
    citecolor=adaptive_blue,
    filecolor=adaptive_blue,
    urlcolor=adaptive_blue,
    }

\titlespacing*{\section}{0pt}{1.0ex plus .2ex minus .2ex}{0.6ex plus .2ex}
\titlespacing*{\subsection}{0pt}{0.8ex plus .2ex minus .2ex}{0.4ex plus .2ex}
\setlist[itemize]{topsep=2pt,itemsep=1pt,parsep=0pt,leftmargin=*}
\setlist[enumerate]{topsep=2pt,itemsep=1pt,parsep=0pt,leftmargin=*}

\usepackage{tikz} 
\definecolor{lightgray}{gray}{0.9}
\input{assets/icons}

\input{assets/lst}
\usepackage{fontawesome5}

\usetikzlibrary{positioning,fit,arrows.meta,calc,shapes.misc}

\title{DS-Lighting: Making Agent Harnesses Explicit for Data-Science Automation}

\author{
 \textbf{Fan Liu\textsuperscript{1}},
 \textbf{Hao Liu\textsuperscript{1}}
\\
 \textsuperscript{1}The Hong Kong University of Science and Technology (Guangzhou)
\\
\texttt{fliu236@connect.hkust-gz.edu.cn}\\
\texttt{liuh@ust.hk} \\
}

\begin{document}
\maketitle

\begin{abstract}
Large Language Model (LLM) agents have shown promise for automating data-science workflows, yet their end-to-end performance depends critically on the agent harness that represents tasks, manages execution state, constrains output artifacts, and provides evaluation feedback. Existing data-science agents often leave this harness implicit, making results difficult to reproduce, compare, and attribute across heterogeneous tasks. We introduce \ourmethod, a unified harness toolkit that makes harness design explicit for data-science automation. \ourmethod decomposes the harness into four reusable layers: data, workflow, execution, and evaluation, and represents diverse agents as executable operator programs that support both predefined pipelines and adaptive search. We further integrate multiple open-source data-science benchmarks into an MLE-Bench-style task format, enabling controlled comparison under a shared task interface, sandboxed runtime, and metric protocol. Experiments across agents, harnesses, models, and ablations show that explicit harness design improves reproducibility, comparability, and reliability, while reducing avoidable system-level failures in end-to-end data-science workflows. Our code is available at \url{https://github.com/usail-hkust/dslighting}
\end{abstract}

\section{Introduction}

Large Language Model (LLM) agents have recently opened a new paradigm for automating data-science workflows~\cite{you2025datawiseagent}. Given a natural-language objective and associated datasets, they can generate code, interact with execution environments, and iteratively refine analytical solutions. Yet their success is not determined by model capability alone~\cite{lou2026autoharness}: end-to-end performance also depends on the \textit{agent harness}, the system layer that manages data context, workflow execution, feedback, and output evaluation. This harness is especially important for data-science tasks, where heterogeneous inputs, long-horizon execution, and structured output requirements can cause failures even when the model produces plausible reasoning or executable code.

\begin{figure}[t]
    \centering
    \includegraphics[width=0.95\linewidth]{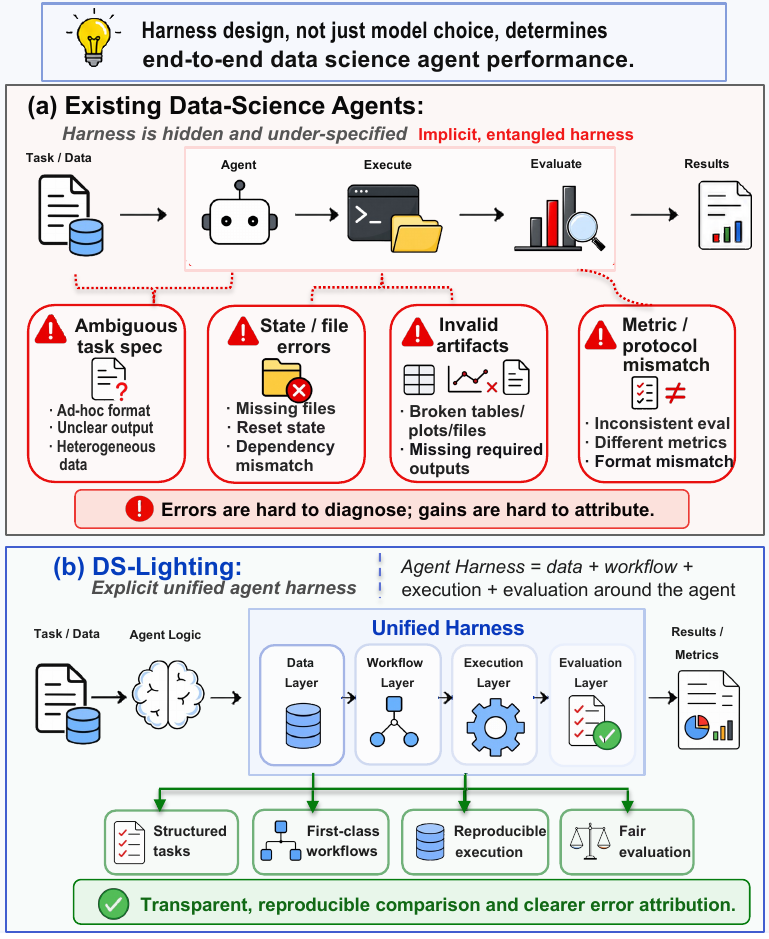}
    \caption{Existing data-science agents vs. \ourmethod{} (ours). The contrast motivates a harness-centered design: existing agents entangle task data, execution, and evaluation, while \ourmethod{} decouples them to ground data, govern execution, and align outputs with metrics.}
    \label{fig:motivation}
    \vspace{-0.3em}
\end{figure}

Despite this progress, existing data-science agents still treat the harness largely as an implementation detail rather than a first-class research object. Prior systems introduce diverse mechanisms for automation: DS-Agent~\cite{guo2024ds} decomposes tasks into a manually designed pipeline for knowledge retrieval, problem analysis, code generation, and execution, while AIDE~\cite{jiang2025aide} frames automation as iterative code optimization over candidate solutions. Other work explores planning~\cite{hong2402data}, self-refinement~\cite{yang2025rdagent}, and multi-step tool use~\cite{zhu2026towardmlmaster}. Yet these systems embed different implicit assumptions about task representation, execution state, intermediate validation, and output evaluation. This makes it difficult to attribute performance differences to stronger models, better workflows, or system-specific harness choices.

Building a data-science harness is a non-trivial systems problem. Unlike answer-only or single-step tool-use tasks~\cite{ning2026codeharness, pan2026natural}, data-science agents must execute long-horizon workflows over real datasets and produce artifacts that can be checked by task-specific metrics. This creates two coupled challenges. First, the harness must unify diverse workflows, from fixed pipelines to adaptive search and multi-agent collaboration, so heterogeneous agents can be instantiated, orchestrated, and compared through a common interface. Second, it must support reliable end-to-end execution by grounding heterogeneous inputs, maintaining consistent code and file states, and enforcing task-specific output constraints. Without these mechanisms, agents may follow plausible reasoning paths yet execute incompatible workflows, solve the wrong problem, corrupt runtime state, or produce artifacts that cannot be evaluated.

To address these challenges, we introduce \ourmethod, a unified harness toolkit that makes harness design explicit for LLM-based data-science automation. \ourmethod decomposes the harness into four reusable layers: a data layer that converts raw tasks and assets into structured task contracts, a workflow layer that represents agents as executable operator programs, an execution layer that provides sandboxed stateful runtime control, and an evaluation layer that validates artifacts and returns standardized feedback. Under a common interface, this design supports both predefined pipelines and adaptive search-based workflows. For controlled evaluation, \ourmethod further integrates multiple open-source data-science benchmarks into an MLE-Bench-style task format, enabling agents and harnesses to be compared under the same task interface, execution environment, and metric protocol.

Our contributions are threefold.
First, we formulate the agent harness as a first-class object in LLM-based data-science automation, showing that task representation, execution state, artifact constraints, and evaluation feedback are key to reliable and comparable end-to-end performance.
Second, we introduce \ourmethod{}, a unified harness toolkit that decomposes data-science automation into four reusable layers: data, workflow, execution, and evaluation. Using an operator-program abstraction, \ourmethod{} represents diverse agents as executable workflows and supports both predefined pipelines and adaptive search under a shared execution interface.
Third, we build a unified evaluation suite by converting multiple open-source data-science benchmarks into an MLE-Bench-style task format, enabling agents and harness frameworks to be compared under the same task interface, sandboxed runtime, and metric protocol. Experiments across agents, harnesses, ablations, models, and failure modes show that explicit harness design improves reproducibility, comparability, and end-to-end reliability.

\section{Related Work}

\textbf{Data Science Benchmarks.}
Recent benchmarks evaluate LLM agents on data-science tasks ranging from exploratory analysis to end-to-end modeling. Data-analysis benchmarks test table understanding, cleaning, transformation, visualization, and iterative analysis~\cite{hu2024infiagent-dabench}; for example, DA-Code~\cite{huang2024code} emphasizes fine-grained code-generation operations, while DSEval~\cite{zhang2024benchmarkingdseval} targets interactive refinement. Data-modeling benchmarks instead require complete machine-learning pipelines, from dataset inspection and feature engineering to training and evaluation~\cite{chan2025mlebench, huang2023mlagentbench}.

\textbf{LLM Data Science Agents.}
LLM-based agents have attracted increasing attention for automating end-to-end data-science workflows~\cite{you2025datawiseagent}. Existing systems generally follow two paradigms: fixed workflows and adaptive search. Fixed-workflow agents decompose tasks into predefined stages such as problem analysis, code generation, execution, and validation~\cite{wang2025dsmentor}; examples include DS-Agent~\cite{guo2024ds}, which uses a procedural pipeline, and Data Interpreter~\cite{hong2402data}, which organizes analysis through hierarchical workflows with incremental validation. Adaptive agents, including AIDE~\cite{jiang2025aide} and AutoKaggle~\cite{li2024autokaggle}, iteratively propose, execute, evaluate, and refine candidate solutions~\cite{ou2025automind, zhang2024aflow}.

\textbf{Agent Harnesses.}
Beyond data science, recent work has highlighted evaluation harnesses, executable environments, and runtime infrastructure for LLM agents. Benchmarks such as WebArena~\cite{zhou2023webarena}, AppWorld~\cite{trivedi2024appworld}, and OSWorld~\cite{xie2024osworld} move beyond static question answering by requiring tool use, stateful interaction, and programmatic validation. Frameworks such as HAL~\cite{kapoor2025holistic} standardize orchestration, logging, cost tracking, trajectory inspection, and reproducible comparison, while AutoHarness~\cite{lou2026autoharness} and SafeHarness~\cite{lin2026safeharness} study harness constraints for reliability and safety. However, these efforts are not designed for data-science workflows, where data context, code execution, output artifacts, and metric-based evaluation must be managed jointly. \ourmethod fills this gap with a data-science-specific harness that separates data, workflow, execution, and evaluation concerns while supporting existing agents and flexible operator-program workflows.

\section{Preliminaries}
\label{sec:preliminaries}

\textbf{Data Science Task.}
A data-science task is represented as,
\begin{equation}
x = (q, D, A),
\end{equation}
where $q$ is the natural-language objective (e.g., ``predict house prices from historical sales records''), $D$ specifies task data assets (e.g., tables, images, time series, graphs, or multimodal inputs), and $A$ provides auxiliary requirements such as documentation, external knowledge, submission constraints, or evaluation metrics. The task output $y$ is the required artifact, such as an analytical report, prediction file, or model.

\textbf{Agent Harness.}
A data-science agent is governed by an \textit{agent harness} $\mathcal{H}$, the system layer that manages how the model uses task context, executes workflows, and incorporates feedback. In \ourmethod, the harness is organized around four responsibilities: constructing data context, instantiating workflows, executing actions, and evaluating artifacts.  Given a task $x$, the agent produces an output according to,
\begin{equation}
y \sim \pi_\theta(\cdot \mid x, \mathcal{H}),
\end{equation}
where $\pi_\theta$ denotes the underlying LLM. 

\textbf{Agent Workflow.}
Within the harness, an agent is represented as a workflow $W$ that specifies how data-science capabilities are organized and executed across multiple steps. A workflow is,
\begin{equation}
W = (\mathcal{O}, C_W, \lambda),
\end{equation}
where $\mathcal{O}$ denotes the operator library of reusable atomic capabilities (e.g., data inspection, code generation, model training, validation, and result submission), $C_W$ denotes the workflow controller that composes and invokes operators during execution (e.g., fixed ordering, conditional branching, iterative refinement, or search-based selection), and $\lambda$ denotes workflow configurations (e.g., prompts, model settings, temperature, maximum iterations, time budget, and validation thresholds).

At step $t$, the selected operator $\operatorname{Op}_t \in \mathcal{O}$ invokes an atomic capability, possibly calling the LLM, interacting with the execution environment, and returning an observation. A simple workflow may follow a fixed sequence (e.g., \textsc{InspectData}, \textsc{GenerateCode}, \textsc{TrainModel}, \textsc{Validate}, and \textsc{Submit}), whereas an adaptive workflow selects operators from intermediate observations, such as retrying code generation after an execution error or switching models after weak validation scores.

\textbf{Execution Loop.}
The workflow and the harness interact through an iterative loop. At step $t$, the state $s_t$ summarizes the current execution context (e.g., task requirements, available data, code context, generated files, and previous feedback). The controller selects an operator $\operatorname{Op}_t=C_W(s_t)$, and the selected operator conditions the model to generate an action,
\begin{equation}
a_t \sim \pi_\theta(\cdot \mid s_t, \operatorname{Op}_t).
\end{equation}
The execution layer runs the action and returns an observation $o_t=\operatorname{Exec}(a_t,s_t)$ (e.g., outputs, errors, artifacts, validation signals, or metric feedback). The harness then updates the state,
\begin{equation}
s_{t+1} = \operatorname{Update}_{\mathcal{H}}(s_t, a_t, o_t).
\end{equation}
Executing $W$ yields a trajectory $\tau_W=\{(s_t,\operatorname{Op}_t,a_t,o_t)\}_{t=1}^{T}$, from which the final output $y$ is produced and evaluated.

\section{DS-Lighting: A Layered Agent Harness}
\label{sec:architecture}

\begin{figure*}[t]
    \centering
    \includegraphics[width=0.95\textwidth]{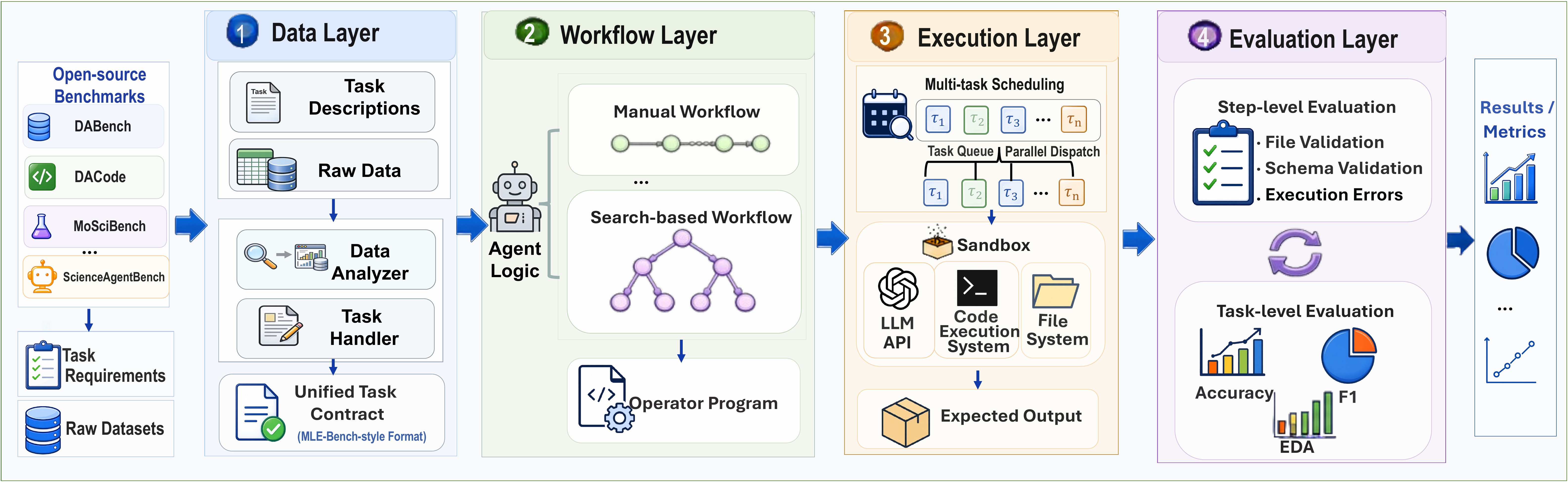}
    \caption{Overview of \ourmethod{} as a layered agent harness. The data layer constructs task contracts, the workflow layer organizes agent capabilities into operator programs, the execution layer controls sandboxed interaction and artifact production, and the evaluation layer validates outputs and returns feedback.}
    \label{fig:method_framework}
    \vspace{-0.9em}
\end{figure*}

\ourmethod realizes the agent harness $\mathcal{H}$ as four coordinated layers that transform raw task inputs into executable workflows and evaluated outputs. The \textit{Data Layer}~\ref{sec:data_layer} builds structured task contracts from task descriptions and data files; the \textit{Workflow Layer}~\ref{sec:workflow_layer} turns each contract into an operator program; the \textit{Execution Layer}~\ref{sec:execution_layer} runs actions in a controlled environment and returns observations; and the \textit{Evaluation Layer}~\ref{sec:evaluation_layer} validates artifacts, step-level feedback, and task-level metrics. Figure~\ref{fig:method_framework} illustrates the framework, while Appendix~\ref{app:usage_examples} provides compact code-style examples for loading tasks, running predefined or custom agents, and launching benchmark suites.

\subsection{Data Layer}
\label{sec:data_layer}

The data layer builds the task context used by downstream workflows. Raw data-science inputs often mix underspecified instructions with heterogeneous assets, making it hard for agents to infer the goal, locate relevant files, understand schemas, and satisfy output requirements. To reduce this ambiguity, the data layer maps the raw task description $q_{\mathrm{raw}}$ and data assets $D_{\mathrm{raw}}$ to a structured task contract $x=(q,D,A)$ through two components: \textit{Data Analyzer} and \textit{Task Handler}. The Data Analyzer summarizes the assets into a compact data report, and the Task Handler merges it with the task description to produce an agent-ready contract.

\textbf{Data Analyzer.}
Data Analyzer provides a data perception stage before workflow execution. Given the raw data assets $D_{\mathrm{raw}}$, it produces an agent-readable report $R_D=\operatorname{DataAnalyzer}(D_{\mathrm{raw}})$ by extracting two types of information: a compact data profile $\phi(D_{\mathrm{raw}})$ and a size-aware sample $S_D \subset D_{\mathrm{raw}}$. The profile summarizes data organization (e.g., directory layout, file formats, schemas, and column-level statistics) and output-format requirements (e.g., column ordering, schema constraints, required file names, and submission fields). The sample provides representative records or file snippets, allowing the agent to inspect actual data content without loading the full dataset. 

\textbf{Task Handler.}
Task Handler converts the natural-language task description and the data report into a standardized task contract, written as $x=\operatorname{TaskHandler}(q_{\mathrm{raw}},R_D)$. The resulting contract instantiates $x=(q,D,A)$ by formalizing the task objective $q$, constructing the standardized data specification $D$, and collecting auxiliary requirements $A$ (e.g., input/output paths, submission schema, evaluation constraints, and expected artifacts). This task contract provides a consistent interface for workflow construction in the next layer. In practice, the data layer converts diverse open-source benchmarks (e.g., DABench, DACode, etc) into a unified MLE-Bench-style task format. Each task includes a description, public inputs, hidden references, sample submissions, and output constraints, enabling consistent execution, artifact validation, and evaluation across heterogeneous benchmarks. Details are provided in Appendix~\ref{app:task_grounding}.

\subsection{Workflow Layer}
\label{sec:workflow_layer}

The workflow layer represents each agent as an executable operator program conditioned on the task contract. Rather than hard-coding monolithic pipelines, \ourmethod separates reusable capabilities from control logic: operators define atomic actions, while a controller schedules them during execution. Fixed workflows use stable operator sequences, whereas search-based workflows select operators dynamically from the current state for exploration, refinement, and recovery. This shared abstraction lets diverse agents be built from extensible components and compared under the same execution and evaluation interface.

In a \textit{manual workflow}, the controller follows a predefined operator sequence, selecting operators by workflow step rather than by the current state,
\begin{equation}
C_W(t)=\operatorname{Op}_t,\quad 
\operatorname{Op}_t \in (\operatorname{Op}_1,\ldots,\operatorname{Op}_K).
\end{equation}
This captures existing agents or user-defined pipelines with fixed procedures (e.g., problem analysis, code generation, execution, and validation) without modifying their internal logic. 

In a \textit{search-based workflow}, the controller selects operators according to the current state:
\begin{equation}
\operatorname{Op}_t = C_W(s_t), \quad \operatorname{Op}_t \in \mathcal{O}.
\end{equation}
Here, $C_W$ may be implemented as a search algorithm or an LLM-based policy, enabling adaptive exploration, refinement, and recovery based on intermediate observations.

At each step, the selected operator $\operatorname{Op}_t$ conditions the model to generate an action
$
a_t \sim \pi_\theta(\cdot \mid s_t, \operatorname{Op}_t),
$
providing the workflow with its next operation. The workflow layer is the development interface for building data-science agents: existing agents can use built-in templates, while new ones extend the operator library, compose workflow programs, or define custom controllers. By separating capabilities, workflow structure, and control policy, \ourmethod unifies reusable templates and flexible operator-program designs, allowing agents to consume grounded task contracts, interact with the runtime, and receive evaluation-aligned feedback through one orchestration interface. Appendix~\ref{app:workflow_layer} provides further details.

\subsection{Execution Layer}
\label{sec:execution_layer}
\vspace{-0.05in}
The execution layer grounds workflow actions into concrete interactions with the environment. Given an action $a_t$ produced by an operator, it executes the action in a sandboxed runtime and returns an observation:
$o_t = \operatorname{Exec}(a_t, s_t),$ where $s_t$ provides the current execution context (e.g., available data files, code context, intermediate variables, generated artifacts, and runtime configuration). The observation $o_t$ may include program outputs, execution errors, generated files, validation signals, or resource usage.  The harness then updates the state $s_{t+1} = \operatorname{Update}_{\mathcal{H}}(s_t, a_t, o_t)$, incorporating the effects of execution such as generated files, variables, logs, and feedback. Over the workflow, this produces a trajectory $\tau = \{(s_t, \operatorname{Op}_t, a_t, o_t)\}_{t=1}^{T}$ from which evaluation and feedback can be derived.

This structure lets each operator run with full context while preserving intermediate results and feedback across steps. For concurrent workflows $\{x_1, x_2, \dots\}$, the execution layer uses a scheduler to coordinate actions under shared resource constraints: cross-task admission selects executable workflows, and within-task dispatch runs ready actions in parallel. Global limits on CPU/GPU allocation and LLM concurrency are enforced throughout, allowing agents to focus on generating actions and consuming observations. In practice, this execution governance tracks code context, dependencies, generated files, variables, logs, and resource usage, keeping long-horizon workflows state-consistent and reducing avoidable failures from missing files, stale states, or invalid runtime assumptions.

\subsection{Evaluation Layer}
\label{sec:evaluation_layer}

The evaluation layer validates outputs and provides structured feedback. It operates at two levels: step-level evaluation, which checks intermediate artifacts during execution, and task-level evaluation, which computes the final score after completion. 

\textbf{Step-Level Evaluation Module.}
Step-level evaluation invokes operators during execution. Given state $s_t$ and evaluation action $a_t$, the evaluator returns an observation,

\begin{equation}
\vspace{-0.3em}
o_t = \operatorname{EvalStep}(a_t, s_t),
\vspace{-0.3em}
\end{equation}

which may include file-structure checks, schema validation, execution diagnostics, missing-output detection, or semantic feedback. These observations are returned in structured, agent-readable form and incorporated into subsequent workflow decisions through the execution loop.

\textbf{Task-Level Evaluation Module.}
Task-level evaluation scores a completed workflow. Given task $x$, trajectory $\tau_W$, and output $y$, the evaluator applies
\begin{equation}
m = G(x, \tau_W, y),
\end{equation}
where $m$ is the task-level score. For data-analysis tasks, $G$ may assess the completeness and correctness of reports, plots, or analytical conclusions. For modeling tasks, $G$ typically computes prediction metrics such as accuracy, AUC, F1, or RMSE.  Together, step-level and task-level evaluation implement evaluation alignment: evaluator-defined requirements are exposed not only as final scores, but also as structured feedback during execution. This allows agents to repair missing artifacts, schema mismatches, execution errors, or invalid submissions before final evaluation.
\section{Experiments}

\begin{table*}[!t]
\centering
\caption{
\textbf{ Benchmark comparison of data-science agents reproduced with the \ourmethod{} all-in-one toolkit.}
The table includes both manual-workflow agents and adaptive/search-workflow agents, demonstrating that \ourmethod{} can instantiate heterogeneous agent workflows under the same task interface, execution environment, and evaluator.
All metrics are accuracy-style scores, where higher values are better ($\uparrow$); \textbf{Overall Avg} reports the mean performance across tasks.
}
\label{tab:data_analysis_llm_blocks}
\vspace{-0.2em}
\renewcommand\tabcolsep{5pt}
\renewcommand\arraystretch{1.02}
\footnotesize

\resizebox{0.95\textwidth}{!}{
\begin{tabular}{l|
c|c|c|c|c|c|c}
\Xhline{1.2pt}

\rowcolor{CadetBlue!20}
\textbf{Task}
& \textbf{AutoKaggle}
& \textbf{Data Interpreter}
& \textbf{DSAgent}
& \textbf{DeepAnalyze}
& \textbf{AIDE}
& \textbf{AutoMind}
& \textbf{ReAct} \\
\rowcolor{CadetBlue!10}
\textbf{Workflow}
& Manual & Manual & Manual & Manual & Adaptive & Adaptive & Adaptive \\
\Xhline{1pt}

\rowcolor{gray!20}
\multicolumn{8}{c}{\llmname{DeepSeek-V3.1-Terminus}} \\

Comprehensive Preprocessing$\uparrow$
& 0.60470 & 0.71110 & 0.75560 & 0.54550 & 0.62220 & 0.70450 & 0.62222 \\

\rowcolor{gray!10}
Correlation Analysis$\uparrow$
& 0.88410 & 0.91550 & 0.88890 & 0.64290 & 0.80560 & 0.86110 & 0.93056 \\

Distribution Analysis$\uparrow$
& 0.85480 & 0.89060 & 0.87500 & 0.70970 & 0.81250 & 0.76560 & 0.89062 \\

\rowcolor{gray!10}
Feature Engineering$\uparrow$
& 0.77550 & 0.86000 & 0.86000 & 0.69390 & 0.80000 & 0.78000 & 0.86000 \\

Machine Learning$\uparrow$
& 0.78950 & 0.89470 & 0.89470 & 0.61110 & 0.94737 & 0.78950 & 0.89474 \\

Outlier Detection$\uparrow$
& 0.73530 & 0.80000 & 0.77140 & 0.51430 & 0.68570 & 0.65710 & 0.71429 \\

\rowcolor{gray!10}
Summary Statistics$\uparrow$
& 0.85060 & 0.88640 & 0.91110 & 0.67820 & 0.82220 & 0.78890 & 0.85556 \\

\rowcolor{lightgray}
\textbf{Overall Avg}
& 0.78493 & 0.85119 & 0.85096 & 0.62794 & 0.78508 & 0.76381 & 0.82400 \\
\Xhline{1pt}

\rowcolor{gray!20}
\multicolumn{8}{c}{\llmname{gpt-5-mini}} \\

Comprehensive Preprocessing$\uparrow$
& 0.71111 & 0.77778 & 0.82222 & 0.64444 & 0.75556 & 0.75556 & 0.68889 \\

\rowcolor{gray!10}
Correlation Analysis$\uparrow$
& 0.86111 & 0.87500 & 0.88889 & 0.75000 & 0.88889 & 0.93056 & 0.84722 \\

Distribution Analysis$\uparrow$
& 0.89062 & 0.85938 & 0.85938 & 0.73438 & 0.90625 & 0.89062 & 0.85938 \\

\rowcolor{gray!10}
Feature Engineering$\uparrow$
& 0.78000 & 0.84000 & 0.84000 & 0.68000 & 0.86000 & 0.86000 & 0.78000 \\

Machine Learning$\uparrow$
& 0.78947 & 0.78947 & 0.73684 & 0.73684 & 0.78947 & 0.84211 & 0.78947 \\

Outlier Detection$\uparrow$
& 0.77143 & 0.77143 & 0.77143 & 0.60000 & 0.77143 & 0.80000 & 0.71429 \\

\rowcolor{gray!10}
Summary Statistics$\uparrow$
& 0.90000 & 0.93333 & 0.93333 & 0.62222 & 0.91111 & 0.91111 & 0.82222 \\

\rowcolor{lightgray}
\textbf{Overall Avg}
& 0.81482 & 0.83520 & 0.83601 & 0.68113 & 0.84039 & 0.85571 & 0.78592 \\

\Xhline{1.2pt}
\end{tabular}}
\vspace{-0.8em}
\end{table*}

We evaluate \ourmethod{} through four harness-level research questions:

\begin{itemize}
    \item \textbf{RQ1: Can \ourmethod{} unify diverse agents through the workflow layer?}
    \item \textbf{RQ2: How does \ourmethod{} compare with existing agent-harness frameworks?}
    \item \textbf{RQ3: Are the key components of \ourmethod{} useful for reliable evaluation and execution?}
    \item \textbf{RQ4: What further insights can be drawn from additional analyses?}
\end{itemize}

\subsection{Experimental Setup}

\textbf{Data-Science Tasks.}
We evaluate \ourmethod{} with a unified, all-in-one toolkit that integrates representative benchmarks, including DABench~\cite{hu2024infiagent-dabench}, DACode~\cite{huang2024code}, MosciBench~\cite{liu2026towards}, and ScienceAgentBench~\cite{chen2025scienceagentbench}. This setup allows different agents and harness configurations to be evaluated through the same task interface, with each benchmark runnable by a single line of code for reproducible large-scale evaluation. The suite spans closed-form data analysis, executable code generation, multimodal scientific discovery, and scientific program generation; detailed task statistics are provided in Appendix~\ref{app:dataset_details}.

\textbf{LLM Agents and Harnesses.}
\textit{LLM Agents.} \ourmethod natively integrates and reproduces seven state-of-the-art data-science agents in an all-in-one toolkit, covering both manual-workflow agents (e.g., AutoKaggle~\cite{li2024autokaggle}, Data Interpreter~\cite{hong2402data}, DS-Agent~\cite{guo2024ds}, DeepAnalyze~\cite{zhang2025deepanalyze}) that follow fixed operator sequences and adaptive/search-workflow agents (e.g., AIDE~\cite{jiang2025aide}, AutoMind~\cite{ou2025automind}, ReAct~\cite{yao2022react}) that iteratively draft, debug, and refine code. All agents are instantiated through the same data, workflow, execution, and evaluation layers and run in a unified sandbox environment with controlled dependencies, ensuring reproducible and directly comparable evaluations.
\textit{Agent Harness Frameworks.} To the best of our knowledge, there is currently no dedicated agent-harness framework tailored to data-science agents, whose workflows require heterogeneous data access, code execution, artifact generation, and benchmark-specific validation. We therefore adapt representative general-purpose agent harnesses to the data-science setting as baselines, including \texttt{VanillaHarness}~\cite{yao2022react}, \texttt{LangChain}~\cite{chase2022langchain}, \texttt{AutoGen}~\cite{wu2024autogen}, and \texttt{OpenHands}~\cite{wang2025openhands}. All harnesses use the same task interface, sandboxed execution environment, and official evaluation metrics, making the comparison focus on harness design rather than differences in task setup.

\textbf{Evaluation Metrics.} We evaluate task performance using the official metric specified by each benchmark. Concretely, these metrics fall into two categories: for benchmark-level tasks, we report an overall accuracy that summarizes task success under the unified evaluation protocol; for Kaggle-style competition tasks, where each task defines its own evaluation objective, we report the corresponding official score (e.g., classification accuracy, F1, RMSE, etc.). To further characterize system efficiency, we additionally report token consumption and wall-clock time. A complete list of metrics is provided in Appendix~\ref{app:dataset_details}.

\textbf{Implementation Details.} For the main agent-comparison experiments, we instantiate each agent following its official implementation and recommended configuration whenever available. Unless otherwise specified, all agents use \llmname{DeepSeek-V3.1-Terminus} as the shared base model, which controls for model choice and makes performance differences more attributable to agent workflows and harness design. We additionally conduct a base-model study by fixing the agent framework and varying the underlying LLM, including \llmname{gpt-5.5}, \llmname{claude-opus-4.7}, \llmname{gpt-5.4-mini}, \llmname{gemini-3.1-pro}, and \llmname{qwen3.5-plus}. For further analyses where the agent framework is not the variable of interest, we use the general-purpose ReAct framework as the default agent framework.
\subsection{Unified Evaluation of Data-Science Agents}
To answer RQ1, Table~\ref{tab:data_analysis_llm_blocks} provides a unified benchmark comparison of seven data-science agents that are uniformly reproduced by \ourmethod{}'s all-in-one toolkit, covering both manual-workflow agents and adaptive/search-workflow agents across seven DABench task types and two backbone LLMs. We summarize the main findings as observations (abbreviated as Obs.). \textbf{Obs.\ding{182} Workflow style matters across task types.} Fixed-workflow agents are particularly reliable on standardized analyses: under \llmname{DeepSeek-V3.1-Terminus}, Data Interpreter reaches 0.91550 on Correlation Analysis and 0.88640 on Summary Statistics, while DSAgent reaches 0.88890 and 0.91110 on the same tasks. These results suggest that predefined workflows remain effective when tasks can be decomposed into stable data-analysis steps. \textbf{Obs.\ding{183} Robustness and peak performance are distinct agent properties.}
Manual-workflow agents like Data Interpreter and DSAgent achieve strong averages, showing stable performance on standardized tasks. In contrast, search-based agents like AIDE, AutoMind, and ReAct often top selected task types, but less uniformly. Thus, unified evaluation should report both aggregate performance and task-level behavior.
Additional data-modeling and efficiency results appear in Appendix~\ref{app:additional_results}.
\label{sec:component_analysis}
\begin{figure*}[t]
    \centering
    \includegraphics[width=0.90\textwidth]{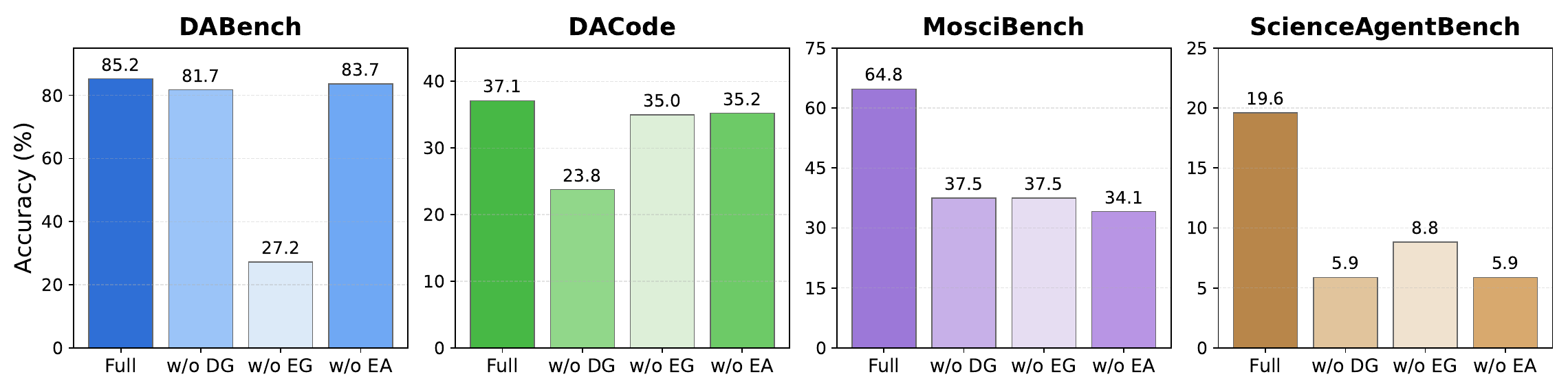}
    \caption{Ablation of \ourmethod{}'s system-level mechanisms. }
    \label{fig:ablation_study_actual_accuracy}
    \vspace{-0.8em}
\end{figure*}

\subsection{Comparison with Agent-Harness Frameworks}

\begin{table}[t]
\centering
\caption{Comparison with existing agent-harness frameworks. Higher values are better.}
\label{tab:harness_framework_comparison}
\vspace{-0.2em}
\small
\setlength{\tabcolsep}{3pt}
\renewcommand{\arraystretch}{1.0}
\resizebox{0.95\linewidth}{!}{
\begin{tabular}{lcccc}
\toprule
\rowcolor{CadetBlue!20}
\textbf{Framework} & \textbf{DABench} & \textbf{DACode} & \textbf{MosciBench} & \textbf{ScienceAgentBench} \\
\midrule
\texttt{VanillaHarness} & \textbf{0.8521} & 0.3073 & 0.4659 & 0.1863 \\
\texttt{LangChain} & 0.8210 & 0.3013 & 0.5000 & 0.1373 \\
\texttt{AutoGen} & 0.7704 & 0.1974 & 0.1932 & 0.0980 \\
\texttt{OpenHands} & 0.7626 & 0.2361 & 0.3182 & 0.0882 \\
\texttt{\ourmethod (ours)} & \textbf{0.8521} & \textbf{0.3709} & \textbf{0.6477} & \textbf{0.1961} \\
\bottomrule
\end{tabular}}
\vspace{-0.5em}
\end{table}

\begin{figure}[t]
    \centering
    \includegraphics[width=0.90\linewidth]{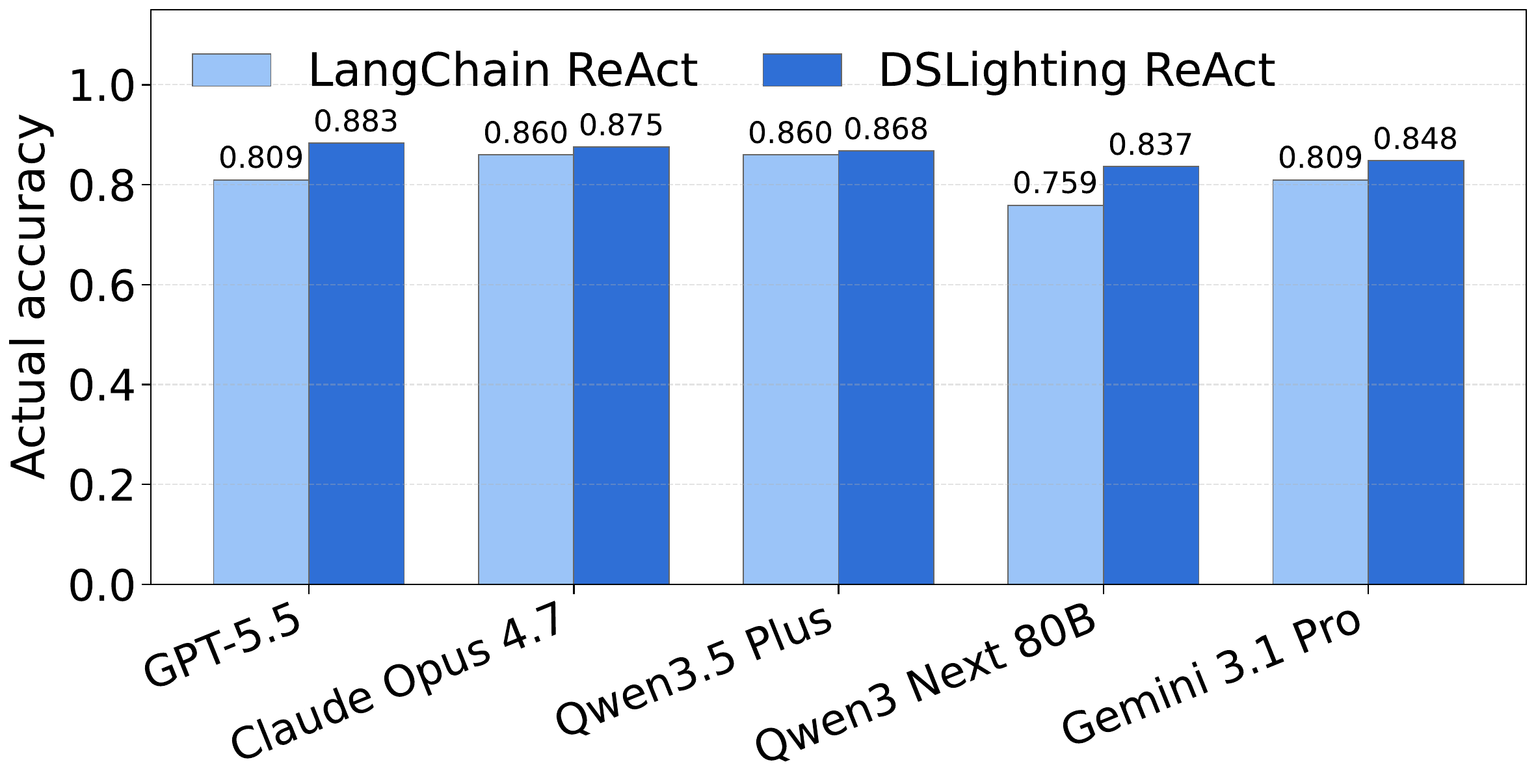}
    \caption{Performance across different models.}
    \label{fig:model_sweep}
    \vspace{-0.2em}
\end{figure}

To answer RQ2, we compare \ourmethod{} with representative agent-harness frameworks under the same task, model, and execution settings. Each framework receives the same public task contract, input files, backbone model, step budget, execution feedback, and final official evaluator. \ourmethod{} never uses hidden labels, private answers, or intermediate metric scores; its extra checks only enforce public output requirements, including artifact existence, filenames, directory structure, and schema consistency. This setup isolates harness-level execution and validation from differences in task information or evaluator access. Table~\ref{tab:harness_framework_comparison} shows that \ourmethod{} is best or tied-best across all four benchmarks, highlighting the value of explicit task context, execution-state management, artifact constraints, and evaluation feedback. \textbf{Obs.\ding{182} General-purpose harnesses are competitive on simple settings but transfer less reliably.}
On DABench, \ourmethod{} matches \texttt{VanillaHarness} (0.8521), indicating that a lightweight harness can be sufficient when tasks are relatively standardized. However, the gap becomes clear on more heterogeneous benchmarks: \ourmethod{} improves over the strongest baseline on DACode (0.3709 vs. 0.3073), MosciBench (0.6477 vs. 0.5000), and ScienceAgentBench (0.1961 vs. 0.1863). This pattern suggests that data-science harness design becomes more important as tasks require richer data access, code execution, and artifact validation. \textbf{Obs.\ding{183} Harness specialization improves robustness across benchmark types.}
General-purpose frameworks show uneven transfer: \texttt{LangChain} is competitive on MosciBench but weaker on ScienceAgentBench, while \texttt{AutoGen} and \texttt{OpenHands} trail across benchmarks. Under the same task interface and metrics, \ourmethod{}'s consistent gains point to the value of structured task contracts, controlled execution, and benchmark-aware evaluation.

\subsection{Ablation Study of System-Level Mechanisms}
To answer RQ3, we ablate the three mechanisms introduced in Section 4: Data Grounding (DG), Execution Governance (EG), and Evaluation Alignment (EA). Each variant removes one mechanism while keeping the benchmark inputs, base agent, execution environment, and final evaluator unchanged. Specifically, w/o DG removes the structured task contract and provides agents with raw task descriptions and files; w/o EG disables stateful sandbox governance and artifact tracking while retaining basic code execution; and w/o EA removes intermediate evaluator-aligned feedback while preserving the final official metric. These controlled variants isolate the effects of each mechanism in Figure~\ref{fig:ablation_study_actual_accuracy}.
\textbf{(1) Data grounding preserves task fidelity.}
Removing DG lowers macro-average accuracy from 0.5167 to 0.3721, with large drops on DACode (0.3709 to 0.2376), MosciBench (0.6477 to 0.3750), and ScienceAgentBench (0.1961 to 0.0588). This confirms that grounding is not merely extra prompt context: it makes files, schemas, fields, and output contracts legible to the agent. \textbf{(2) Execution governance is the main reliability bottleneck.} Removing EG causes the largest decline, reducing macro-average accuracy to 0.2714 and collapsing DABench from 0.8521 to 0.2724. Thus, task understanding must be paired with runtime observation, artifact validation, and recovery to keep code execution, file state, and submissions synchronized. \textbf{(3) Evaluation alignment enables task-relevant repair.} Removing EA lowers macro-average accuracy to 0.3971, especially on MosciBench (0.6477 to 0.3409) and ScienceAgentBench (0.1961 to 0.0588), showing that evaluator-aligned feedback turns grading failures into actionable repair signals.

\begin{figure}[t]
    \centering
    \includegraphics[width=0.85\linewidth]{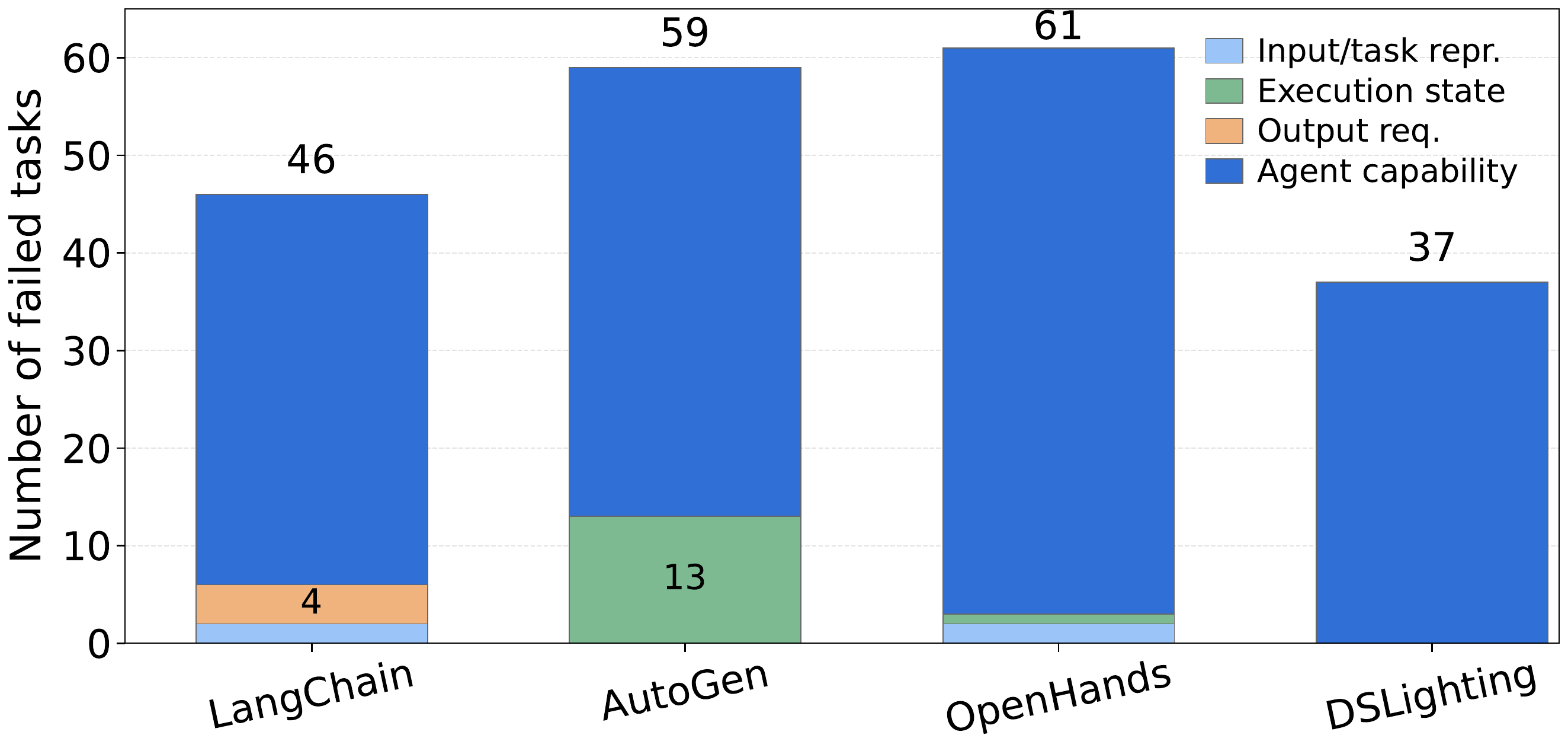}
    \caption{Failure-mode analysis.}
    \label{fig:failure_modes}
    \vspace{-0.2em}
\end{figure}

\subsection{Further Analysis}

To answer RQ4, we analyze model sensitivity and failure modes on DABench. \textbf{Obs.\ding{182} Harness design can outweigh backbone scale.} Figure~\ref{fig:model_sweep} shows that \ourmethod{} paired with Qwen surpasses LangChain paired with GPT-5.5. This suggests that end-to-end performance is not determined by model strength alone: a well-structured harness can better expose task context, regulate execution, and convert model outputs into valid artifacts.
\textbf{Obs.\ding{183} The harness mainly removes avoidable system-level failures.} To better understand where agents fail, we use an LLM-as-judge protocol to inspect failed trajectories and categorize each error by its dominant cause, such as task-grounding mistakes, execution-control failures, evaluation-alignment issues, or intrinsic model limitations. Figure~\ref{fig:failure_modes} shows that \ourmethod{} reduces non-capability failures, especially those tied to grounding, runtime control, and evaluator-aligned repair. As these avoidable failures decrease, the remaining errors concentrate more on intrinsic model limitations, suggesting that the harness improves reliability by removing system-level failure sources rather than merely masking reasoning errors.

\section{Conclusion}
We presented \ourmethod, a unified harness toolkit for developing and evaluating data-science agents. By making task context, workflow execution, and evaluation explicit through reusable data, workflow, execution, and evaluation layers, \ourmethod supports heterogeneous agents under a common interface, enables controlled and reproducible comparison, and reveals failures in long-horizon workflows. Experiments show that harness design is central to agent performance: \ourmethod improves robustness across agents and benchmarks, reduces avoidable execution and evaluation failures, and helps smaller backbone models approach stronger ones.

\newpage
\section*{Limitations}

Although \ourmethod{} is evaluated on diverse data-science benchmarks, it does not cover the full range of real-world industrial workflows, such as private databases, streaming data, human-in-the-loop review, or domain-specific governance. \ourmethod{} also depends on executable environments, external libraries, and benchmark-provided evaluators; despite our unified runtime and shared evaluation protocol, results may still vary with dependency versions, resource limits, API availability, or sandbox configuration. Future work will extend \ourmethod{} to more realistic deployments and improve environment standardization, logging, and portability.

\bibliography{custom}
\clearpage
\appendix
\input{sections/appendix}

\end{document}

%% file: assets/icons.tex
\definecolor{icontext}{RGB}{255, 255, 255} 

\definecolor{datateal}{RGB}{0, 128, 128}     
\definecolor{datablue}{RGB}{30, 144, 255}    
\definecolor{datagreen}{RGB}{46, 139, 87}    
\definecolor{datapurple}{RGB}{138, 43, 226}  

\definecolor{datacyan}{RGB}{0, 139, 139}     
\definecolor{datamidblue}{RGB}{70, 130, 180} 
\definecolor{dataviolet}{RGB}{106, 90, 205}  
\definecolor{datasea}{RGB}{32, 178, 170}     

\definecolor{probred}{RGB}{178, 34, 34}      
\definecolor{proborange}{RGB}{255, 140, 0}   
\definecolor{probpurple}{RGB}{128, 0, 128}   
\definecolor{probnavy}{RGB}{0, 0, 128}       
\definecolor{probforest}{RGB}{34, 139, 34}   
\definecolor{probcyan}{RGB}{0, 139, 139}     
\definecolor{probblack}{RGB}{0, 0, 0}        



%% file: assets/lst.tex
\tcbuselibrary{listings}

\lstdefinestyle{pythonstyle}{
    language=Python,
    backgroundcolor=\color{gray!5},
    basicstyle=\ttfamily\footnotesize,
    keywordstyle=\color{blue},
    commentstyle=\color{green!50!black},
    stringstyle=\color{red!60!black},
    numberstyle=\tiny\color{gray},
    breakatwhitespace=false,
    breaklines=true,
    captionpos=b,
    keepspaces=true,
    numbers=right,
    numbersep=5pt,
    showspaces=false,
    showstringspaces=false,
    showtabs=false,
    tabsize=4,
    frame=none,
    rulecolor=\color{black},
    framerule=0.5pt
}

\newtcblisting{pythoncode}{
    listing only,
    listing options={style=pythonstyle},
    enhanced,
    arc=2mm,
    outer arc=1mm,
    colback=gray!5,
    colframe=gray!50!black,
    boxrule=0.2mm
}

%% file: sections/appendix.tex
\newtcolorbox{AIBoxNoTitle}{
  enhanced,
  breakable,
  colback=gray!3,
  colframe=gray!45,
  boxrule=0.5pt,
  arc=1.5pt,
  left=2pt,
  right=2pt,
  top=2pt,
  bottom=2pt
}

\section{Usage Examples}
\label{app:usage_examples}

\ourmethod{} exposes a compact user-facing API for moving from a single task to full benchmark evaluation under the same task contract. At the task level, users can call \texttt{load\_data} to construct the public data context and then invoke \texttt{run\_agent} to execute a selected workflow. At the agent level, predefined workflows can be swapped by changing only the \texttt{workflow} argument, while custom workflows can be defined by composing asynchronous operators inside a \texttt{BaseWorkflow}. The same configuration can then be passed to \texttt{DSBenchmark} for suite-level evaluation. Figures~\ref{fig:dslighting_single_task_usage}--\ref{fig:dslighting_benchmark_usage} illustrate this progression from single-task execution, to predefined and custom agents, to benchmark runs.

\begin{figure}[H]
\begin{AIBoxNoTitle}
{\tiny
\begin{lstlisting}[language=Python]
from dslighting import load_data, run_agent

task = load_data("bike-sharing-demand")
result = run_agent(data=task, workflow="aide", model="gpt-4o")

# Or launch the same task directly by id.
result = run_agent(task_id="bike-sharing-demand", workflow="aide", model="gpt-4o")
\end{lstlisting}}
\end{AIBoxNoTitle}
\caption{Running a \ourmethod{} agent on a single task.}
\label{fig:dslighting_single_task_usage}
\vspace{-0.4em}
\end{figure}

Figure~\ref{fig:dslighting_single_task_usage} shows the shortest path for running one data-science task: the loader builds the task contract, and the agent runner handles workflow execution and artifact collection. This keeps the task interface identical whether the input is an already loaded task object or a benchmark task id.

\begin{figure}[H]
\begin{AIBoxNoTitle}
{\tiny
\begin{lstlisting}[language=Python]
from dslighting import Agent, run_agent

agent = Agent(workflow="aide", model="gpt-4o")
result = agent.run(task_id="bike-sharing-demand")

# Switch workflows without changing the task interface.
result = run_agent(task_id="bike-sharing-demand", workflow="react", model="gpt-4o")
\end{lstlisting}}
\end{AIBoxNoTitle}
\caption{Using and switching predefined \ourmethod{} agent workflows. Supported workflows include \texttt{aide}, \texttt{react}, \texttt{dsagent}, \texttt{automind}, \texttt{autokaggle}, \texttt{data\_interpreter}, \texttt{deepanalyze}, and \texttt{aflow}.}
\label{fig:dslighting_predefined_agent_usage}
\vspace{-0.4em}
\end{figure}

Figure~\ref{fig:dslighting_predefined_agent_usage} illustrates that predefined agents share the same public interface. In practice, users can compare workflows such as \texttt{aide}, \texttt{react}, \texttt{dsagent}, or \texttt{autokaggle} by changing one argument while keeping the model, task id, and benchmark runner fixed.

\begin{figure}[H]
\vspace{-0.55em}
\begin{AIBoxNoTitle}
{\tiny
\begin{lstlisting}[language=Python]
from dslighting.arch.operators import Operator
from dslighting.arch.workflows import BaseWorkflow
class MyOperator(Operator):
    async def __call__(self, description, data_dir):
        return {"answer": f"Solved task from {data_dir}"}
class MyAgent(BaseWorkflow):
    async def solve(self, description, io_instructions, data_dir, output_path):
        result = await self.operators["solve"](description, data_dir)
        output_path.write_text(result["answer"])
agent = MyAgent(operators={"solve": MyOperator()}, services={}, agent_config={})
result = agent.run(data="data/bike-sharing-demand", task="Analyze the dataset")
\end{lstlisting}}
\end{AIBoxNoTitle}
\vspace{-0.7em}
\caption{Defining and running a minimal custom \ourmethod{} agent.}
\label{fig:dslighting_custom_agent_usage}
\vspace{-0.9em}
\end{figure}

Custom agents follow the same contract but expose the workflow internals. As shown in Figure~\ref{fig:dslighting_custom_agent_usage}, each workflow implements \texttt{solve(...)}, calls operators, and writes the required artifact to \texttt{output\_path}, making custom logic compatible with the same evaluator.

\begin{figure}[H]
\begin{AIBoxNoTitle}
{\tiny
\begin{lstlisting}[language=Python]
from dslighting.api import DSBenchmark
from dslighting.core import ConfigBuilder

config = ConfigBuilder().build_config(workflow="aide", model="gpt-4o")
result = DSBenchmark("dabench", data_dir="data/dabench").run(config=config)

# benchmark_id can be "dabench", "dacode", "moscibench", or "sciencebench".
result = DSBenchmark(benchmark_id, data_dir="data/benchmark-data").run(config=config)
\end{lstlisting}}
\end{AIBoxNoTitle}
\caption{Running benchmark suites with \ourmethod{}. Internally, \textsc{ScienceAgentBench} is selected with the benchmark id \texttt{sciencebench}.}
\label{fig:dslighting_benchmark_usage}
\vspace{-0.4em}
\end{figure}

Finally, Figure~\ref{fig:dslighting_benchmark_usage} shows how the single-task setup scales to benchmark-level evaluation. The benchmark runner reuses the same workflow configuration across all tasks in a suite and applies the corresponding official evaluation protocol.

\section{Task Grounding and Benchmark Normalization}
\label{app:task_grounding}

To evaluate heterogeneous data-science benchmarks through a common interface, \ourmethod{} normalizes every task into an MLE-Bench-style competition directory. Each task contains a natural-language description, preparation and grading scripts, public agent-visible artifacts, and private references reserved for evaluation:

\begin{lstlisting}[language={},caption={Normalized task layout.}]
task-id/
  config.yaml
  description.md
  prepare.py
  grade.py
  raw/
  prepared/
    public/       # visible to the agent
    private/      # hidden references for grading
\end{lstlisting}

The conversion assigns each asset a clear role. \texttt{raw/} stores original benchmark files; \texttt{prepare.py} materializes \texttt{prepared/public/} and \texttt{prepared/private/}; and \texttt{grade.py} implements the task-local evaluator. The accompanying metadata specifies the task type, description, preparer, grader, and expected submission contract:

\begin{lstlisting}[language={},caption={Example normalized task metadata.}]
competition_type: code
description: description.md
grader:
  name: compare_csv
  grade_fn: file:grade.py:grade
preparer: file:prepare.py:prepare
\end{lstlisting}

At runtime, the Task Handler reads this metadata and constructs the task contract $x=(q,D,A)$. Here, $q$ contains the objective and metric description; $D$ exposes only \texttt{prepared/public/} plus Data Analyzer summaries such as file types, schemas, missing-value statistics, and sampled rows; and $A$ defines the required output artifact, including filename, format, file/directory kind, sample submission, and grading mode. Agents are sandboxed with access only to public artifacts and must write outputs according to this contract. After execution, \ourmethod{} validates the artifact path and format, then routes evaluation through the unified service: artifact-submission tasks call the local \texttt{grade.py}, while open-ended or judge-based tasks use the judge evaluator. This normalization prevents private-answer leakage, removes benchmark-specific I/O assumptions from agents, and enables all workflows to be compared through the same task contract and evaluation path.

\section{Workflow Layer and Operator Programs}
\label{app:workflow_layer}

\ourmethod{} represents each agent as an executable \emph{operator program} conditioned on the task contract $x=(q,D,A)$. The contract specifies the task objective, public data artifacts, output constraints, and evaluation interface, while the workflow defines how an agent acts on this contract over time. This separation keeps the task interface fixed and makes agents differ only in their control logic.

Formally, \ourmethod{} maintains an operator library $\mathcal{O}=\{\mathrm{Op}_1,\ldots,\mathrm{Op}_K\}$, where each operator is an atomic executable capability such as task analysis, planning, code generation, code execution, artifact validation, review, or memory summarization. At step $t$, the workflow state $s_t$ stores the task contract, messages, generated code, execution outputs, validation results, artifacts, and workflow-specific memory. A controller $C_W$ selects an operator and applies it to the current state:
\begin{equation}
    \mathrm{Op}_t = C_W(s_t,x), \qquad \mathrm{Op}_t \in \mathcal{O}.
\end{equation}
The selected operator conditions the model or tool executor to produce the next action, and the returned observation updates the workflow state:
\begin{align}
    a_t &\sim \pi_\theta(\cdot \mid x,s_t,\mathrm{Op}_t), \\
    s_{t+1} &= U_W(s_t,\mathrm{Op}_t,a_t,y_t),
\end{align}
where $y_t$ is the observation from the environment or evaluator, such as generated code, sandbox output, validation feedback, or a score estimate.

\paragraph{Manual workflows.}
A manual workflow follows a predefined operator schedule, selecting operators by step index rather than adaptive feedback:
\begin{equation}
    \mathrm{Op}_t=C_W(t)=\mathrm{Op}_{\sigma_W(t)},
    \qquad \mathrm{Op}_{\sigma_W(t)}\in\mathcal{O}.
\end{equation}
Here, $\sigma_W$ is the fixed sequence specified by the workflow, such as \texttt{analyze} $\rightarrow$ \texttt{plan} $\rightarrow$ \texttt{generate-code} $\rightarrow$ \texttt{execute} $\rightarrow$ \texttt{validate}. This captures existing agents and user-defined pipelines by wrapping each step as an operator with a common input/output interface.

\paragraph{Search-based workflows.}
A search-based workflow chooses operators from the current state. The controller may be a heuristic, tree-search procedure, or LLM-based policy, enabling the workflow to branch, refine attempts, recover from execution errors, revisit earlier decisions, or terminate once the artifact satisfies the output contract. For example, failed execution can trigger debugging, a valid but weak result can trigger refinement, and successful validation can trigger submission.

\paragraph{Execution interface.}
Manual and search-based workflows share the same harness interface: each operator receives the grounded task contract and returns structured outputs such as plans, code, logs, reviews, or artifact paths. Side effects are isolated in the sandbox workspace, and final artifacts are checked against the output contract before grading. Thus, \ourmethod{} can orchestrate heterogeneous agents while comparing them under the same grounding, runtime, validation, and benchmark evaluation pipeline.

\section{Dataset Details}
\label{app:dataset_details}

We evaluate agents on data science benchmarks covering general data analysis, code generation, multimodal scientific discovery, and scientific program generation. The full benchmark suite contains 947 tasks, including 257 from DABench, 500 from DACode, 88 from MoSciBench, and 102 from ScienceAgentBench.

\textbf{DABench.} DABench~\cite{hu2024infiagent-dabench} is a data analysis benchmark designed to evaluate whether language agents can solve closed-form analytical questions over tabular datasets. The benchmark covers common data analysis skills, including summary statistics, correlation analysis, distribution analysis, feature engineering, data preprocessing, outlier detection, and basic machine learning. In our evaluation, DABench contains 257 normalized tasks. Each task requires the agent to inspect the provided data, perform the necessary analysis, and submit an answer in a structured CSV format. Tasks are evaluated automatically using exact-match or tolerance-based grading against ground-truth answers. 

\textbf{DACode.} DACode is a data science code generation benchmark for evaluating agents on practical data manipulation, modeling, and visualization tasks. Unlike closed-form question answering benchmarks,
  DACode emphasizes executable code generation and artifact production. The benchmark includes 500 tasks spanning data wrangling, data interpretation, text and CSV processing, plotting,
  database manipulation, and machine learning tasks such as classification, regression, clustering, and competition-style prediction. Outputs include CSV files, JSON files, database files,
  plots, and machine learning prediction files, which are evaluated using task-specific automatic graders.

\textbf{MoSciBench.} MoSciBench is a multimodal scientific discovery benchmark that evaluates agents on end-to-end scientific analysis tasks. It contains 88 tasks across six scientific domains: Earth science, biomedical engineering, cheminformatics, health psychology, population genomics, and Earth system science. The benchmark covers multiple data modalities, including time series, tabular data, images, mass spectra, molecular structures, genotype matrices, and text. Tasks are organized around scientific discovery question types such as descriptive analysis, correlational reasoning, causal inference, predictive modeling, and pattern discovery. Each task requires the agent to analyze raw scientific data and produce a structured answer that can be automatically evaluated.

\textbf{ScienceAgentBench.} ScienceAgentBench is a benchmark for assessing language agents on data-driven scientific discovery tasks derived from peer-reviewed research. It contains 102 tasks extracted from 44 papers
  across four scientific disciplines: bioinformatics, chemoinformatics, geoscience, and cognitive neuroscience. Each task is formulated as a self-contained scientific program generation
  problem, where the agent must write and execute code to produce the required output artifact. Outputs include figures, tables, arrays, text files, JSON files, and other scientific artifacts.
  Evaluation is performed using task-specific metrics such as image similarity, accuracy, RMSE, MAE, F1, ROC-AUC, exact match, and correlation-based scores.

\textbf{MLE-bench.} additional also test kaggle style competiiton from mle-bench~\cite{chan2025mlebench}. Table~\ref{tab:mlebench_dataset_stats} summarizes the datasets used for data modeling, which are curated to represent heterogeneous predictive tasks across multiple modalities. The benchmark includes tabular and time-series forecasting problems (e.g., \textit{bike-sharing-demand}, \textit{ili}, \textit{demand-forecasting-kernels-only}), classification tasks over sequential signals (e.g., \textit{handwriting}, \textit{liverpool-ion-switching}), and an image-based medical classification task (i.e., \textit{histopathologic-cancer-detection}). It also includes multimodal or weakly structured settings such as \textit{see-click-predict-fix}, which combines spatiotemporal and textual signals for multi-target regression. Each dataset is evaluated using its official competition metric (e.g., RMSLE, RMSE, AUC-ROC, Macro-F1), enabling consistent comparisons of end-to-end modeling quality across agents.

\begin{table}[t!]
\centering
\scriptsize
\setlength{\tabcolsep}{2.0pt}
\renewcommand{\arraystretch}{1.08}
\caption{Kaggle-style competition dataset summary.}
\label{tab:mlebench_dataset_stats}
\begin{adjustbox}{max width=\linewidth}
\begin{tabular}{p{2.15cm} p{1.55cm} p{1.75cm} p{1.15cm}}
\toprule
\rowcolor{CadetBlue!20}
\textbf{Dataset} & \textbf{Modality} & \textbf{Task} & \textbf{Metric} \\
\midrule

bike-sharing-demand
& Tabular + time
& Forecasting (regression)
& RMSLE \\

\rowcolor{gray!10}
handwriting
& Time series
& Multiclass classification
& Accuracy \\

histopathologic-cancer-detection
& Image
& Binary classification
& AUC-ROC \\

\rowcolor{gray!10}
ili
& Time series
& Forecasting (multivariate)
& MSE/MAE \\

demand-forecasting-kernels-only
& Time series
& Forecasting (regression)
& SMAPE \\

\rowcolor{gray!10}
instant-gratification
& Tabular
& Binary classification
& AUC-ROC \\

see-click-predict-fix
& Spatiotemporal + text
& Multi-target regression
& RMSLE \\

\rowcolor{gray!10}
playground-series-s3e1
& Tabular
& Regression
& RMSE \\

playground-series-s3e11
& Tabular
& Regression
& RMSE \\

\bottomrule
\end{tabular}
\end{adjustbox}
\end{table}

\section{Implementation Details}
\label{app:implementation_details}

\paragraph{Unified reproduction of agent-harness frameworks.}
We reproduce VanillaHarness, LangChain, AutoGen, OpenHands, and \ourmethod under the same data-science benchmark adapter to ensure a controlled and easy-to-follow comparison. For each task, the adapter builds a unified task contract containing the original task description, public input files, sample artifacts when available, required output filename or directory structure, and the official evaluation interface, while never exposing private answers, hidden labels, or official scores during solving. All frameworks receive this same contract in the prompt, run inside an isolated per-task workspace with the same public files, model, step budget, and execution feedback, and are graded only after termination using the same official benchmark metric. The general-purpose baselines are adapted to this contract through their native execution mechanisms: VanillaHarness uses a direct ReAct-style code-execution loop, LangChain uses its ReAct message-history interface, AutoGen uses AssistantAgent with PythonCodeExecutionTool, and OpenHands uses its default CLI agent. \ourmethod  follows the same public task interface and final grader, but additionally applies its built-in harness mechanisms for protocol checking, context management, execution governance, and public output-contract inspection. These checks only enforce syntactic public requirements such as artifact existence, filename, and location; they do not reveal correctness, labels, or metric values. Therefore, the comparison isolates framework-level execution behavior under identical task information and evaluation metrics, with \ourmethod 's gains attributable to its runtime design rather than access to extra benchmark supervision.

\paragraph{Operator implementation and contract validation.}
\ourmethod{} does not treat workflow steps as free-form prompt fragments. Instead, each step is implemented as an operator with an explicit asynchronous API, declared inputs and outputs, and controlled side effects. Operators can be composed sequentially, in parallel, conditionally, or through a dynamic DAG runtime, where each edge explicitly binds an upstream output key to a downstream input key. A node is dispatched only after all dependencies have succeeded and all required inputs are available; Figure~\ref{fig:operator_node_example} shows a representative declarative operator node. Interface contracts are enforced at three levels: structured LLM outputs are validated against schemas before downstream use; execution operators check runtime arguments and run generated code inside an isolated sandbox; and benchmark-facing artifacts are checked against the required submission contract, including output name, artifact type, required files, suffix constraints, and format-level requirements such as CSV columns or sample-submission consistency. Figure~\ref{fig:operator_output_example} gives an example of a structured operator output validated before use. Thus, workflow state is passed through explicit typed objects and node results rather than implicit prompt history alone.

\begin{figure*}[htbp]
\begin{AIBoxNoTitle}
{\scriptsize
\begin{lstlisting}
{
  "node_id": "execute_code",
  "op_type": "sandbox",
  "operator_name": "ExecuteAndTestOperator",
  "depends_on": ["generate_code"],
  "input_bindings": [
    {
      "input_key": "code",
      "source_node_id": "generate_code",
      "source_key": "code"
    }
  ],
  "payload": {
    "kwargs": {
      "mode": "script"
    }
  },
  "max_retries": 1,
  "timeout_seconds": 300
}
\end{lstlisting}}
\end{AIBoxNoTitle}
\caption{Example declarative operator node in the dynamic DAG runtime.}
\label{fig:operator_node_example}
\end{figure*}

\begin{figure*}[htbp]
\begin{AIBoxNoTitle}
{\scriptsize
\begin{lstlisting}
{
  "operator_name": "LLMBasedReviewOperator",
  "output_type": "ReviewResult",
  "schema": {
    "is_buggy": "bool",
    "summary": "string",
    "metric_value": "float | null",
    "lower_is_better": "bool | null"
  },
  "example_output": {
    "is_buggy": false,
    "summary": "The code executed successfully and produced a valid submission.",
    "metric_value": 0.5876,
    "lower_is_better": true
  }
}
\end{lstlisting}}
\end{AIBoxNoTitle}
\caption{Example structured operator output validated before downstream use.}
\label{fig:operator_output_example}
\end{figure*}

\section{Additional Experimental Results}
\label{app:additional_results}

\begin{figure}[t]
    \centering
    \includegraphics[width=0.95\linewidth]{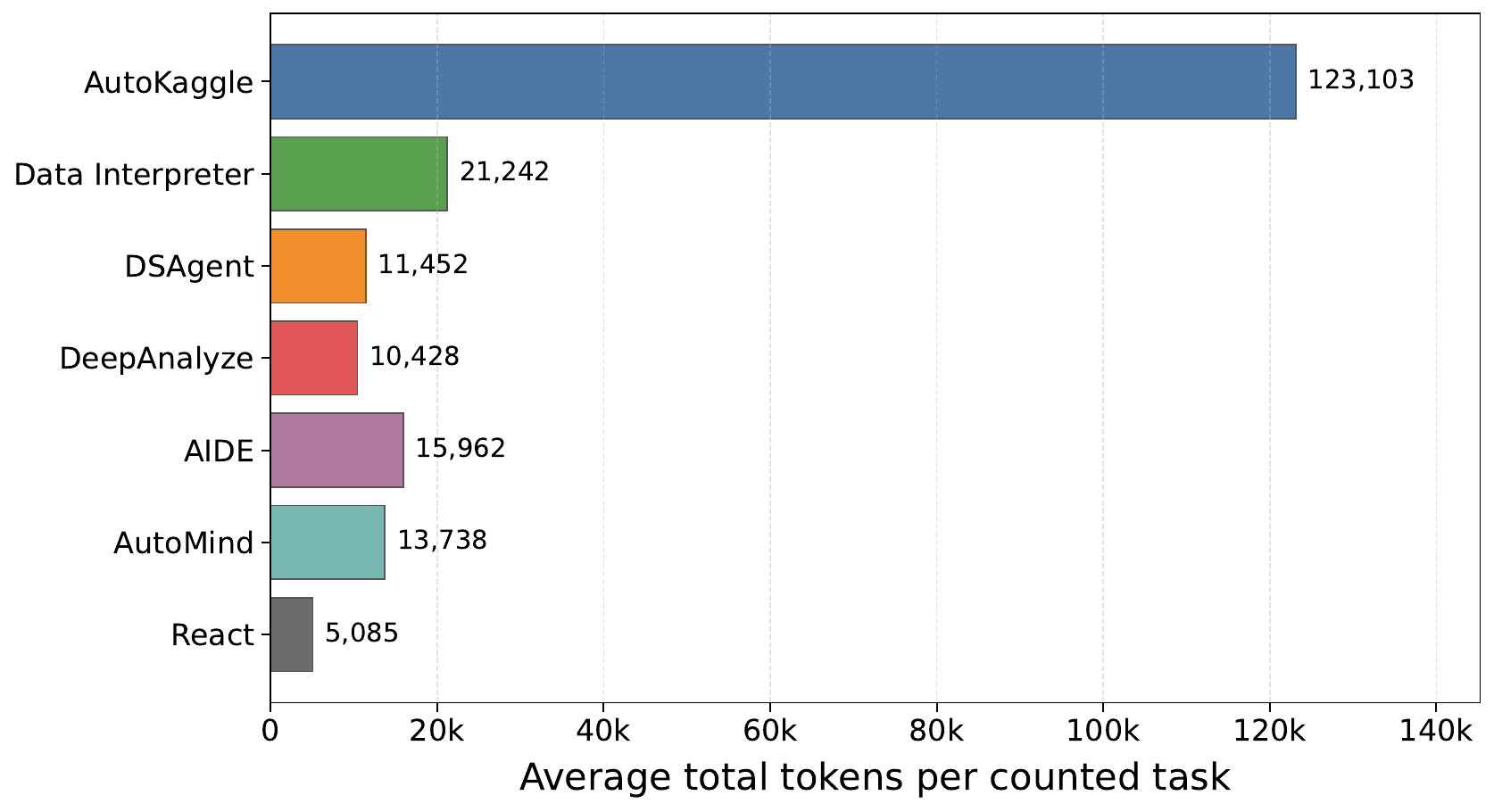}
    \caption{Average token consumption of different agents.}
    \label{fig:da_token_deepseek}
    \vspace{-0.3em}
\end{figure}

\begin{figure}[t]
    \centering
    \includegraphics[width=0.95\linewidth]{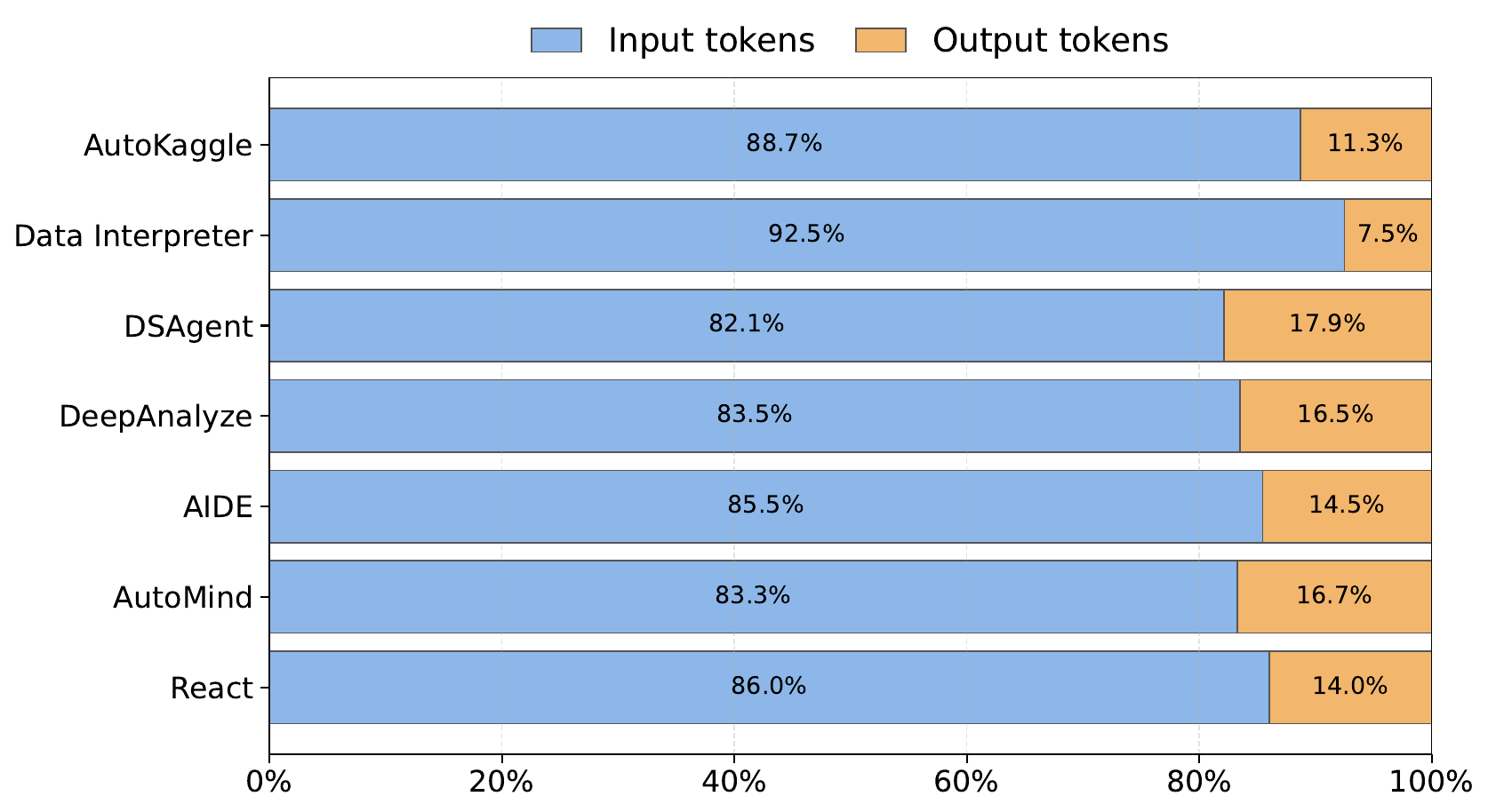}
    \caption{Input/output token composition of different agents.}
    \label{fig:da_token_composition}
    \vspace{-0.3em}
\end{figure}

  \textbf{Token Consumption across Agents.}
 Figure~\ref{fig:da_token_deepseek} compares average total token consumption on \textsc{DABench}, normalized by counted task for a fairer comparison. \textbf{Obs.\ding{182} Token efficiency differs sharply across agents.} React is the most token-efficient agent, using only 5.1K tokens per task on average. By contrast, AutoKaggle consumes 123.1K tokens per task, or 24.21$\times$ more than React. Even relative to the most efficient traditional agent, DeepAnalyze, React reduces token consumption by 51.2\%. This gap reflects workflow design: AutoKaggle maintains larger contexts and repeats reasoning, code generation, and execution cycles, whereas React uses shorter action-observation loops with less context accumulation.
\textbf{Obs.\ding{183} Input context dominates token cost.} Figure~\ref{fig:da_token_composition} decomposes total usage into input and output tokens. Across agents, input tokens account for 82.1\%--92.5\% of the total budget; React follows the same trend, with 86.0\% input tokens and 14.0\% output tokens. Thus, token cost is driven mainly by prompt-side context, including task descriptions, intermediate observations, execution histories, and tool feedback. Reducing redundant context and controlling history growth are therefore key to improving token efficiency in data-analysis agents.

\begin{figure}[t]
    \centering
    \includegraphics[width=0.95\linewidth]{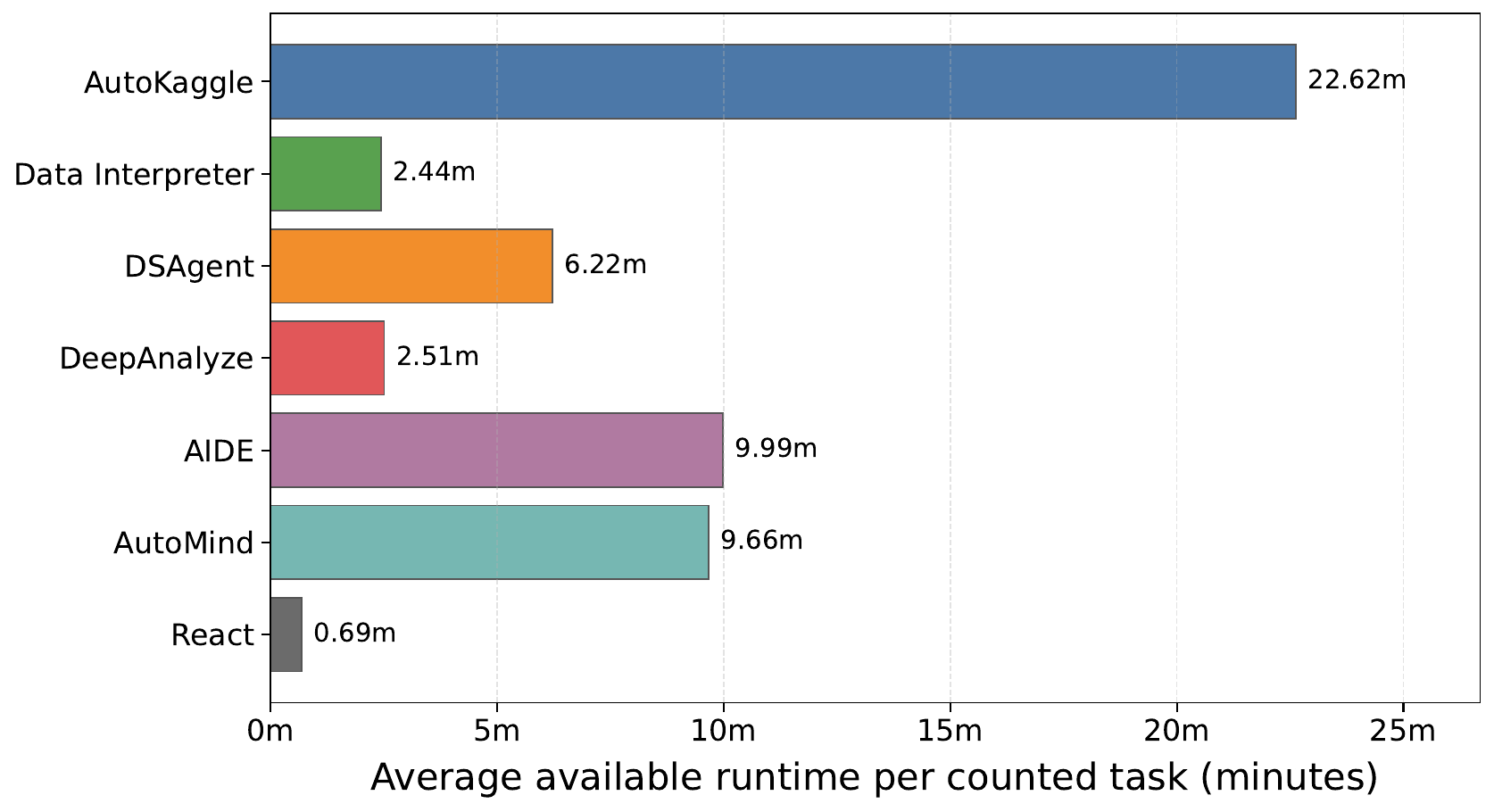}
    \caption{Average running time of different agents.}
    \label{fig:a_time_deepseek}
    \vspace{-0.3em}
\end{figure}

\textbf{Runtime Comparison across Agents.} Figure~\ref{fig:a_time_deepseek} compares average runtime across agent frameworks reproduced under \ourmethod{} on \textsc{DABench}, normalized by counted task for a fairer comparison. \textbf{Obs.\ding{184} Runtime efficiency varies widely across frameworks.} React is the fastest, requiring only 0.69 minutes per task, while AutoKaggle is the slowest, requiring 22.62 minutes per task. This yields a 32.93$\times$ gap between the fastest and slowest frameworks. Among non-React agents, Data Interpreter and DeepAnalyze are the fastest (2.44 and 2.51 minutes per task), DSAgent is moderate (6.22 minutes), and AIDE and AutoMind are slower (9.99 and 9.66 minutes). AutoKaggle is a clear outlier, reaching 141.39 cumulative hours in total.
\textbf{Obs.\ding{185} Workflow complexity drives runtime cost.} Lightweight interpreter-style agents such as Data Interpreter and DeepAnalyze use shorter analysis and code-execution loops, leading to lower overhead. By contrast, search- or competition-oriented frameworks such as AutoKaggle, AIDE, and AutoMind introduce more planning, candidate generation, evaluation, and refinement, increasing both model-call latency and tool-execution cost. Overall, the reproduced agents show a clear trade-off: compact workflows are faster, while multi-stage search and refinement workflows incur much higher cumulative runtime.

\begin{figure}[t]
    \centering
    \includegraphics[width=0.95\linewidth]{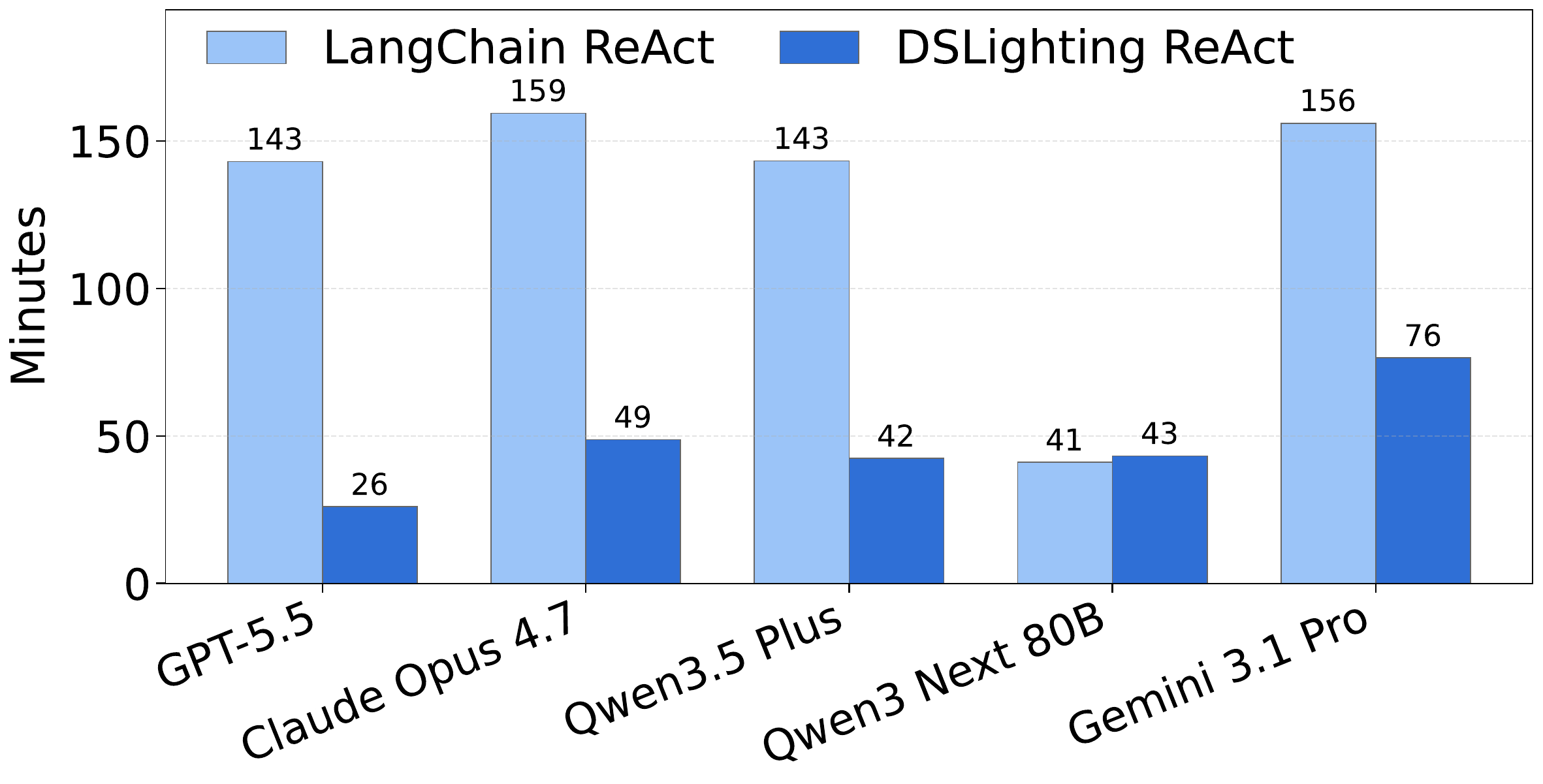}
    \caption{Runtime comparison between LangChain ReAct and \ourmethod{} ReAct across models on \textsc{DABench}.}
    \label{fig:model_sweep_time}
    \vspace{-0.3em}
\end{figure}

\textbf{Runtime Comparison across Frameworks.} Figure~\ref{fig:model_sweep_time} compares LangChain ReAct and \ourmethod{} ReAct across foundation models on \textsc{DABench}. \textbf{Obs.\ding{186} The harness implementation strongly affects runtime.} Across five complete model runs, LangChain ReAct takes 642.7 minutes in total, whereas \ourmethod{} ReAct takes 236.9 minutes, giving a 2.71$\times$ overall speedup. The speedup appears on four of five models, with the largest gain on GPT-5.5 (5.49$\times$), followed by Qwen3.5 Plus (3.38$\times$), Claude Opus 4.7 (3.27$\times$), and Gemini 3.1 Pro (2.04$\times$). \textbf{Obs.\ding{187} Runtime must be interpreted with task completion.} Qwen3 Next 80B is the only case where LangChain ReAct appears faster, but this is mainly due to frequent execution errors and earlier termination rather than more efficient problem solving. Overall, \ourmethod{} ReAct reduces framework-level overhead through a more compact execution loop, showing that runtime is shaped by both the backbone model and the surrounding harness.

\begin{figure}[t]
    \centering
    \includegraphics[width=0.95\linewidth]{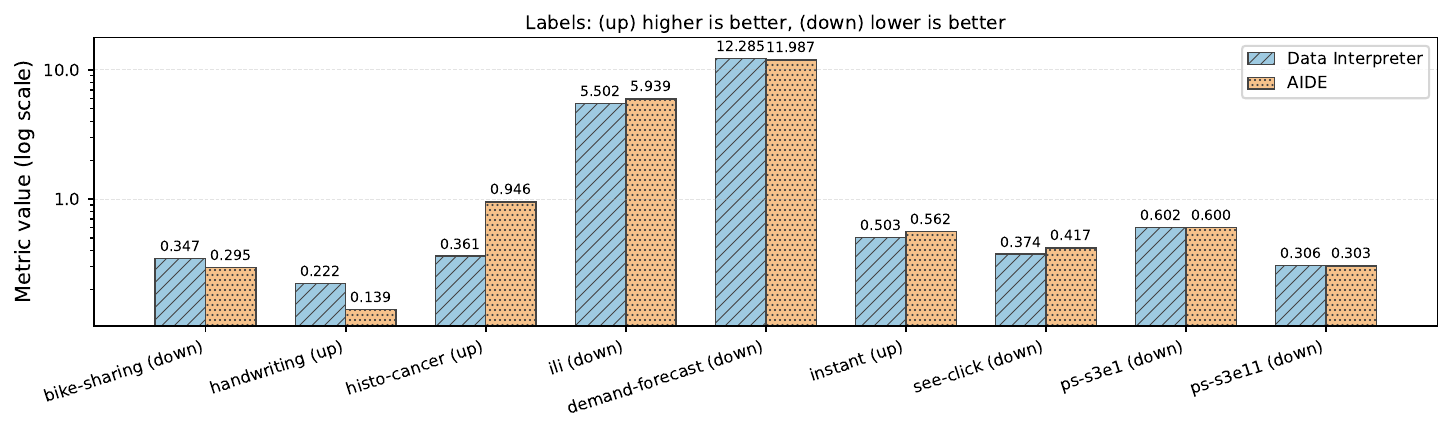}
    \caption{Comparison between Data Interpreter and AIDE on nine Kaggle-style competition datasets with valid results from both agents.}
    \label{fig:di_aide_common}
    \vspace{-0.3em}
\end{figure}

\textbf{Kaggle-Style Competition Tasks.} Figure~\ref{fig:di_aide_common} compares Data Interpreter and AIDE on the nine Kaggle-style datasets where both agents produce valid results. \textbf{Obs.\ding{188} Performance is task-dependent across competition settings.} AIDE performs better on five datasets, while Data Interpreter performs better on four. AIDE is stronger on classification-oriented tasks such as \textit{histopathologic-cancer-detection} and \textit{instant-gratification}, whereas Data Interpreter remains competitive on several forecasting and regression tasks. This suggests that AIDE's iterative search-and-refinement workflow helps on some tasks, but the simpler code-interpreter workflow can be more stable on others; no single framework dominates across all competition-style settings.

\section{Prompt and Task-Contract Examples}
\label{app:prompt_examples}

Figure~\ref{fig:task_contract_schema} and Figure~\ref{fig:task_contract_prompt} illustrate how \ourmethod{} represents each benchmark instance as a structured task contract and renders it into an agent-facing prompt. The contract follows $x=(q,D,A)$, where $q$ specifies the task objective, $D$ describes public data artifacts and analyzer summaries, and $A$ specifies the required answer artifact and official evaluation interface. All compared harnesses receive the same rendered task contract and public data-analysis feedback.

\begin{figure*}[htbp]
\begin{AIBoxNoTitle}
{\scriptsize
\begin{lstlisting}
{
  "task_id": "string",
  "benchmark": "string",
  "q": {
    "task_description": "string",
    "objective": "string",
    "metric": "string or null",
    "metric_direction": "higher_is_better | lower_is_better | null"
  },
  "D": {
    "working_directory": "./",
    "public_artifacts": [
      {
        "path": "string",
        "role": "train | test | sample_submission | metadata | other",
        "modality": "tabular | text | image | time_series | other",
        "format": "csv | json | parquet | txt | png | ...",
        "profile": {
          "status": "ok | degraded | error",
          "rows_sampled": "integer or null",
          "columns_detected": "integer or null",
          "columns": [
            {
              "name": "string",
              "dtype": "string",
              "missing_pct": "number",
              "cardinality": "integer or null",
              "sample_values": ["..."]
            }
          ]
        }
      }
    ]
  },
  "A": {
    "submission": {
      "root_kind": "file | directory",
      "output_name": "string",
      "format": "csv | json | txt | directory | ...",
      "sample_artifact": "string or null"
    },
    "evaluation": {
      "mode": "artifact_submission | judge",
      "official_metric": "string",
      "objective": "higher_is_better | lower_is_better | exact_match | null"
    }
  }
}
\end{lstlisting}}
\end{AIBoxNoTitle}
\caption{Schema of the structured task contract used by \ourmethod{}.}
\label{fig:task_contract_schema}
\end{figure*}

\begin{figure*}[htbp]
\begin{AIBoxNoTitle}
{\scriptsize
\begin{lstlisting}
Task Description:
Compute the requested statistic from the provided Titanic table and write the final
answer to the required submission file.

Data Analyzer Report:
- Artifact: train.csv
  Format: csv
  Modality: tabular
  Rows sampled: 891
  Columns detected: 12
  Columns:
    PassengerId: int64, missing=0.0%, examples=[1, 2, 3]
    Fare: float64, missing=0.0%, examples=[7.25, 71.2833, 8.05]

- Artifact: sample_submission.csv
  Format: csv
  Role: sample submission
  Columns:
    answer: float64, examples=[0.0]

I/O Requirements:
You MUST save the final submission file in the current working directory.
The required output filename is: submission.csv.
Use sample_submission.csv as the schema reference when needed.
Do not rename the final output file.

Evaluation:
The produced artifact will be evaluated by the benchmark's official grader.
\end{lstlisting}}
\end{AIBoxNoTitle}
\caption{Prompt-style rendering of a DABench-style task contract.}
\label{fig:task_contract_prompt}
\end{figure*}